\documentclass[11pt,a4paper]{article}
\usepackage[hyperref]{acl2017}
\usepackage{times}
\usepackage{latexsym}

\usepackage{url}

\usepackage{graphicx}
\usepackage{amssymb,amsmath,bm}
\usepackage{textcomp}
\usepackage{multirow}
\usepackage{dsfont}
\usepackage{color}
\usepackage[normalem]{ulem}
\usepackage{rotating}
\usepackage{booktabs}
\usepackage{makecell}
\usepackage{comment}
\usepackage{amsmath,amsfonts,amssymb}
\usepackage{tikz}
\usepackage{verbatim}
\usepackage{tabularx}
\usepackage{soul}
\usepackage{algorithm}
\usepackage[noend]{algpseudocode}
\usepackage{graphicx}
\usepackage{caption}
\usepackage{subcaption}

\aclfinalcopy 

\title{Sub-domain Modelling for Dialogue Management \\ with Hierarchical Reinforcement Learning}

\author{Pawe\l~Budzianowski, Stefan~Ultes, Pei-Hao~Su, Nikola~Mrk{\v s}i{\' c}, \\ \textbf{Tsung-Hsien~Wen, I\~nigo~Casanueva, Lina~Rojas-Barahona and Milica~Ga{\v s}i{\' c}} \\
  Cambridge University, Engineering Department, Trumpington Street, Cambridge, UK \\
  {\tt \{pfb30,mg436\}@cam.ac.uk} \\
}
\date{}

\begin{document}
\maketitle
\begin{abstract}
Human conversation is inherently complex, often spanning many different topics/domains. This makes policy learning for dialogue systems very challenging. Standard flat reinforcement learning methods do not provide an efficient framework for modelling such dialogues. In this paper, we focus on the under-explored problem of multi-domain dialogue management. First, we propose a new method for hierarchical reinforcement learning using the \emph{option} framework. Next, we show that the proposed architecture learns faster and arrives at a better policy than the existing flat ones do. Moreover, we show how pretrained policies can be adapted to more complex systems with an additional set of new actions. In doing that, we show that our approach has the potential to facilitate policy optimisation for more sophisticated multi-domain dialogue systems.
\end{abstract}

\section{Introduction}
The statistical approach to dialogue modelling has proven to be an effective way of building conversational agents capable of providing required information to the user \cite{POMDP_williams, young2013pomdp}. Spoken dialogue systems (SDS) usually consist of various statistical components, dialogue management being the central one. Optimising dialogue management can be seen as a planning problem and is normally tackled using reinforcement learning (RL). Many approaches to policy management over single domains have been proposed over the last years with ability to learn from scratch 
\cite{fatemi2016policy,GPRL,su2016continuously,williams2016end}.

The goal of this work is to propose a coherent framework for a system capable of managing conversations over multiple dialogue domains. Recently, a number of frameworks were proposed for handling multi-domain dialogue as multiple independent single-domain sub-dialogues \cite{lison2011multi,wang2014policy,Mrksic15,gavsic2015policy}. \newcite{cuayahuitl2016deep} proposed a network of deep Q-networks with an SVM classifier for \emph{domain selection}.
However, such frameworks do not scale to modelling complex conversations over large state/action spaces, as they do not facilitate conditional training over multiple domains. This inhibits their performance, as domains often share sub-tasks where decisions in one domain influence learning in the other ones. 

In this paper, we apply \emph{hierarchical reinforcement learning} (HRL) \cite{barto2003recent} to dialogue management over complex dialogue domains. Our system learns how to handle complex dialogues by learning a multi-domain policy over different domains that operate on independent time-scales with temporally-extended actions. 

HRL gives a principled way for learning policies over complex problems. It overcomes the curse of dimensionality which plagues the majority of complex tasks by reducing them to a sequence of sub-tasks. It also provides a learning framework for managing those sub-tasks at the same time  \cite{dietterich2000hierarchical, sutton1999between,bacon2017}. 

Even though the first work on  HRL dates back to the 1970s, its usefulness for dialogue management is relatively under-explored. A notable exception is the work of Cuay{\'a}huitl \citeyearpar{cuayahuitl2009hierarchical, cuayahuitl2010evaluation},
whose method is based on the MAXQ algorithm \cite{dietterich2000hierarchical} making use of hierarchical abstract machines \cite{parr1998reinforcement}. 
The main limitation of this work comes from the tabular approach which prevents the efficient approximation of the state space and the objective function. This is crucial for scalability of spoken dialogue systems to more complex scenarios.
Parallel to our work, \citet{peng2017composite} proposed another HRL approach, using deep Q-networks as an approximator. In separate work, we found deep Q-networks to be unstable \cite{eddy2017learning};
in this work, we focus on more robust estimators.

The contributions of this paper are threefold. First, we adapt and validate the option framework \cite{sutton1999between} for a multi-domain dialogue system.
Second, we demonstrate that  hierarchical learning for dialogue systems works well with function approximation using the GPSARSA algorithm. We chose the Gaussian process as the function approximator as it provides uncertainty estimates which can be used to speed up  learning and achieve more robust performance.
Third, we show that independently pre-trained domains can be easily integrated into the system and adapted to handle more complex conversations. 

\section{Hierarchical Reinforcement Learning}
\label{sec:3}
Dialogue management can be seen as a control problem: it estimates a distribution over possible user requests -- \emph{belief states}, and chooses what to say back to the user, i.e.~which \emph{actions} to take to maximise positive user feedback -- the \emph{reward}. 
\paragraph{Reinforcement Learning}
The framework described above can be analyzed from the perspective of the \emph{Markov Decision Process} (MDP). 
We can apply RL to our problem where we parametrize an optimal policy $\pi : \mathcal{B} \times \mathcal{A} \rightarrow [0,1]$. 
The learning procedure can either directly look for the optimal policy \cite{sutton1999policy}  
or model the $Q$-value function \cite{RL}:
$$Q^{\pi}(\mathbf{b}, a) = \text{E}_{\pi}  \{ \sum_{k=0}^{T-t} \gamma^k r_{t+k} | \mathbf{b}_t = \mathbf{b}, a_t = a \},$$ 
where $r_t$ is the reward at time $t$ and $0 < \gamma  \leq 1 $ is the discount factor.
Both approaches proved to be an effective and robust way of training dialogue systems online in interaction with real users \cite{milica_real_users, williams2016end}.
\paragraph{Gaussian Processes in RL}
Gaussian Process RL (GPRL) is one of the state-of-the-art
RL algorithms for dialogue modelling \cite{GPRL} where the
$Q$-value function is approximated using Gaussian processes with a zero mean and chosen kernel function $k(\cdot, \cdot)$, i.e.
$$Q(\mathbf{b},a) \sim \mathcal{GP}\left(0, k((\mathbf{b},a),(\mathbf{b},a)) \right).$$ 
\begin{figure}[t!]
\centerline{\includegraphics[width=80mm]{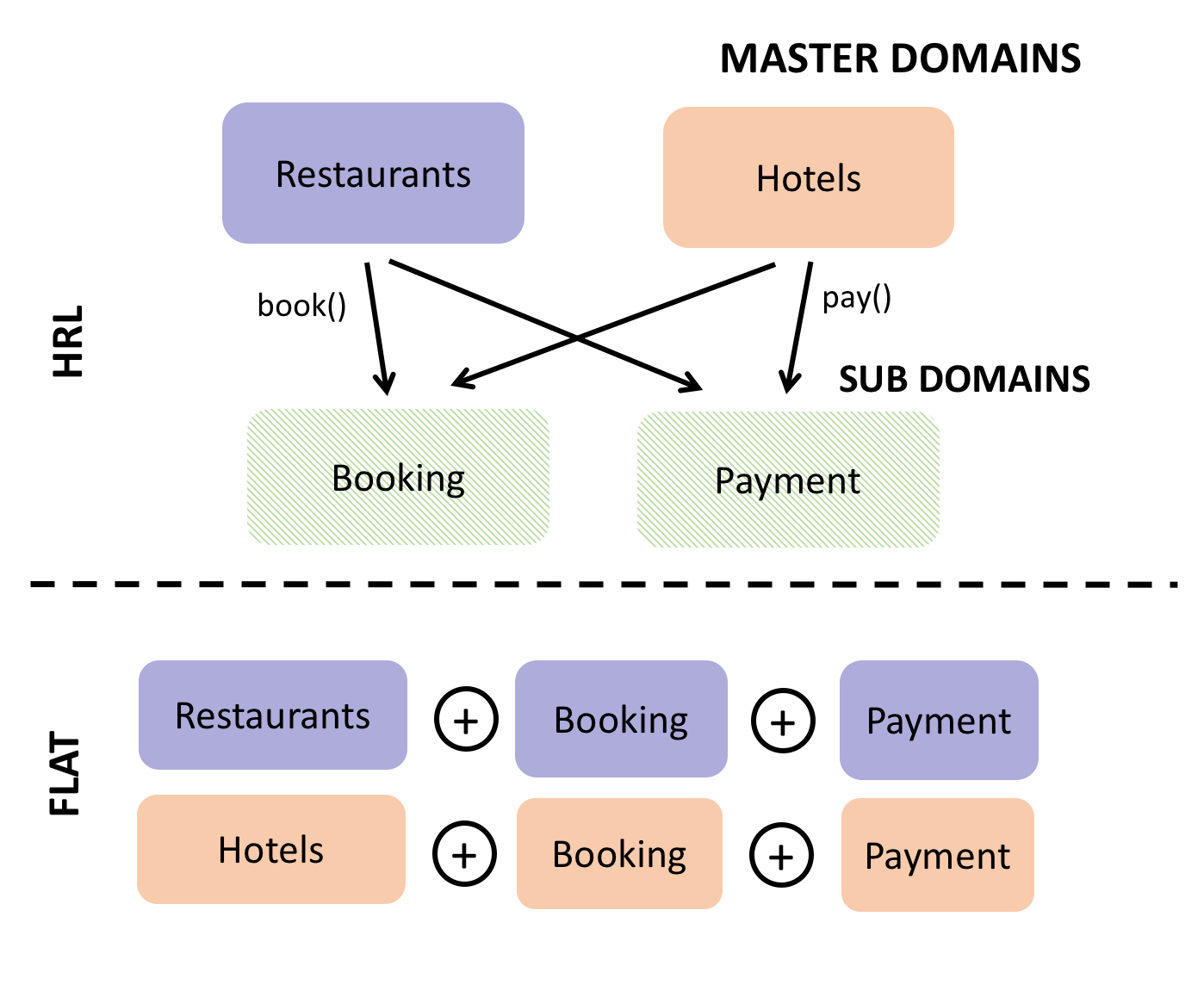}}
\caption{Comparison of two analysed architectures. 
}
\label{fig:model}
\end{figure}
Gaussian processes follow a pure Bayesian framework, which allows one to obtain the posterior given a new collected pair $(\mathbf{b},a)$. The trade-off between exploration and exploitation is handled naturally as given belief state $\mathbf{b}$ at the time $t$ we can sample from posterior $Q(\mathbf{b},a)$ over set of available actions $\mathcal{A}$ to choose the action with the highest sampled Q-value.
GPRL has enabled systems to be trained from scratch in on-line interaction with real users \cite{milica_real_users}.

\paragraph{Hierarchical Policy}
Standard \emph{flat} models where a single Markov Decision Process is responsible for solving multi-task problems have proven to be inefficient. These models have trouble overcoming the cold start problem and/or suffer from the curse of dimensionality \cite{barto2003recent}.
This pattern was also observed with state-of-the-art models proposed recently \cite{mnih2013playing, duan2016benchmarking}. 

To overcome this issue, many frameworks have been proposed in the literature \cite{fikes1972learning, laird1986chunking, parr1998reinforcement}. They make use of hierarchical control architectures and learning algorithms whereby specifying a hierarchy of tasks 
and reusing parts of the state space across many sub-tasks can greatly improve both learning speed and agent performance. 

The key idea is the notion of {\it temporal abstraction} \cite{sutton1999between} where decisions at the given level are not required at each step but can call temporally-extended sub-tasks with their own policies. 

\paragraph{The Option Framework}
One of the most natural generalisations of flat RL methods to complex tasks 
and easily interchangeable with primitive actions
is the \emph{option} model \cite{sutton1999between}.
The option is a generalisation of a single-step action that might span across more than one time-step and can be used as a standard action. From mathematical perspective option is a tuple $\langle \pi, \beta, \mathcal{I} \rangle$ that consists of policy $\pi:  \mathcal{S} \times \mathcal{A} \rightarrow [0,1]$ which conducts the option, stochastic termination condition $\beta : \mathcal{S} \rightarrow [0, 1]$ and an input set $\mathcal{I} \subseteq \mathcal{S}$ which specifies when the option is available.

As we consider hierarchical architectures with temporally extended activities, we have to generalise the MDP to the semi-Markov Decision Process (SMDP) \cite{parr1998reinforcement} where actions can take a variable amount of time to complete. This creates a division between \emph{primitive} actions that span over only one action (and can be seen as a classic reinforcement learning approach)
 and \emph{composite} actions (options) that involve an execution of a sequence of primitive actions. This introduces a policy $\mu$ over options that selects option $o$ in state $s$ with probability $\mu(s,o)$, $o'$s policy might in turn select other options until $o$ terminates and so on. 
The value function for option policies can be defined in terms of the value functions of the flat Markov policies \cite{sutton1999between}:
$$V^{\pi}(s) = \mathbb{E}\{ r_{t+1} + \gamma r_{t+2} +... | E(\pi, s,t)\},$$
where $ E(\pi, s,t)$ is the event of $\pi$ being initiated at time $t$ in $s$. This means we can apply RL methods in HRL using different time-scales.

\begin{algorithm}
\label{alg:euclid}
\footnotesize 
\caption{Hierarchical GPRL}
\begin{algorithmic}[1]
\State Initialize dictionary sets $\mathcal{D}_{\mathcal{M}}, \mathcal{D}_{\mathcal{S}}$ and policies  $\pi_{\mathcal{M}}, \pi_{\mathcal{S}}$ for master and sub-domains accordingly
\For{episode=1:N} 
\State Start dialogue and obtain initial state $\mathbf{b}$
\While{$\mathbf{b}$ is not terminal}
\State Choose action $a$ according to $\pi_m$
\If{$a$ is primitive} 
\State Execute $a$ and obtain next state $\mathbf{b'}$
\State Obtain extrinsic reward $r_e$
\Else
\State Switch to chosen sub-domain
\While{$\mathbf{b}$ is not terminal \textbf{or} $a$ terminates }
\State Choose action $a$ according to $\pi_s$
\State Obtain next state $\mathbf{b'}$
\State Obtain intrinsic reward $r_i$
\State Store transition in $\mathcal{D}_s$
\State $\mathbf{b} \leftarrow\mathbf{b'}$
\EndWhile
\EndIf
\State Store transition in $\mathcal{D}_m$
\State $\mathbf{b} \leftarrow\mathbf{b'}$
\EndWhile
\State Update parameters with $\mathcal{D}_m, \mathcal{D}_s$
\EndFor
\end{algorithmic}
\end{algorithm}

\section{Hierarchical Policy Management}
We propose a multi-domain dialogue system with a pre-imposed hierarchy  that uses the option framework for learning an optimal policy. 
The user starts a conversation in one of the {\it master} domains and switches to the other domains (having satisfied his/her goal) that are seen by the model as {\it sub-domains}. To model individual policies, we can use any RL algorithm. 
In separate work, we found deep RL models performing worse in noisy environment \cite{eddy2017learning}. Thus, we employ
the GPSARSA model from section \ref{sec:3} which proves to handle efficiently noise in the environment. 
The system is trained from scratch where it has to learn appropriate policy using both primitive and temporally extended actions.

We consider two task-oriented master domains providing restaurant and hotel information for the Cambridge (UK) area. Having found the desired entity, the user can then book it for a specified amount of time or pay for it. The two domains have a set of primitive actions (such as \emph{request, confirm} or \emph{inform} \cite{ultes2017pydial}) and a set of composite actions (e.g.,~\emph{book}, \emph{pay}) which call sub-domains shared between them.

The \verb|Booking| and \verb|Payment| domains were created in a similar fashion: the user wants to reserve a table in a restaurant or a room in a hotel for a specific amount of money or duration of time.
The system's role is to determine whether it is possible to make the requested booking. 
The sub-domains operates only on primitive actions and it's learnt following standard RL framework. 

Figure \ref{fig:model} shows the analysed architecture: the \verb|Booking| and \verb|Payment| tasks/sub-domains are shared between two master domains. This means we can train general policies for those sub-tasks that adapt to the current dialogue given the information passed to them by the master domains. 

Learning proceeds on two different time-scales. Following \cite{dietterich2000hierarchical, kulkarni2016hierarchical}, we 
use pseudo-rewards to train sub-domains using an 
internal critic which assesses whether the sub-goal has been reached. 

The master domains are trained using the reward signal from the environment. If a one-step option (i.e.,~a primitive action) is chosen, we obtain immediate extrinsic reward while for the composite actions the master domain waits until the sub-domain terminates and the cumulative reward information is passed back to the master domain. The pseudo-code for the learning algorithm is given in Algorithm $1$.    
  
  \begin{figure}[t!]
\centerline{\includegraphics[width=\linewidth]{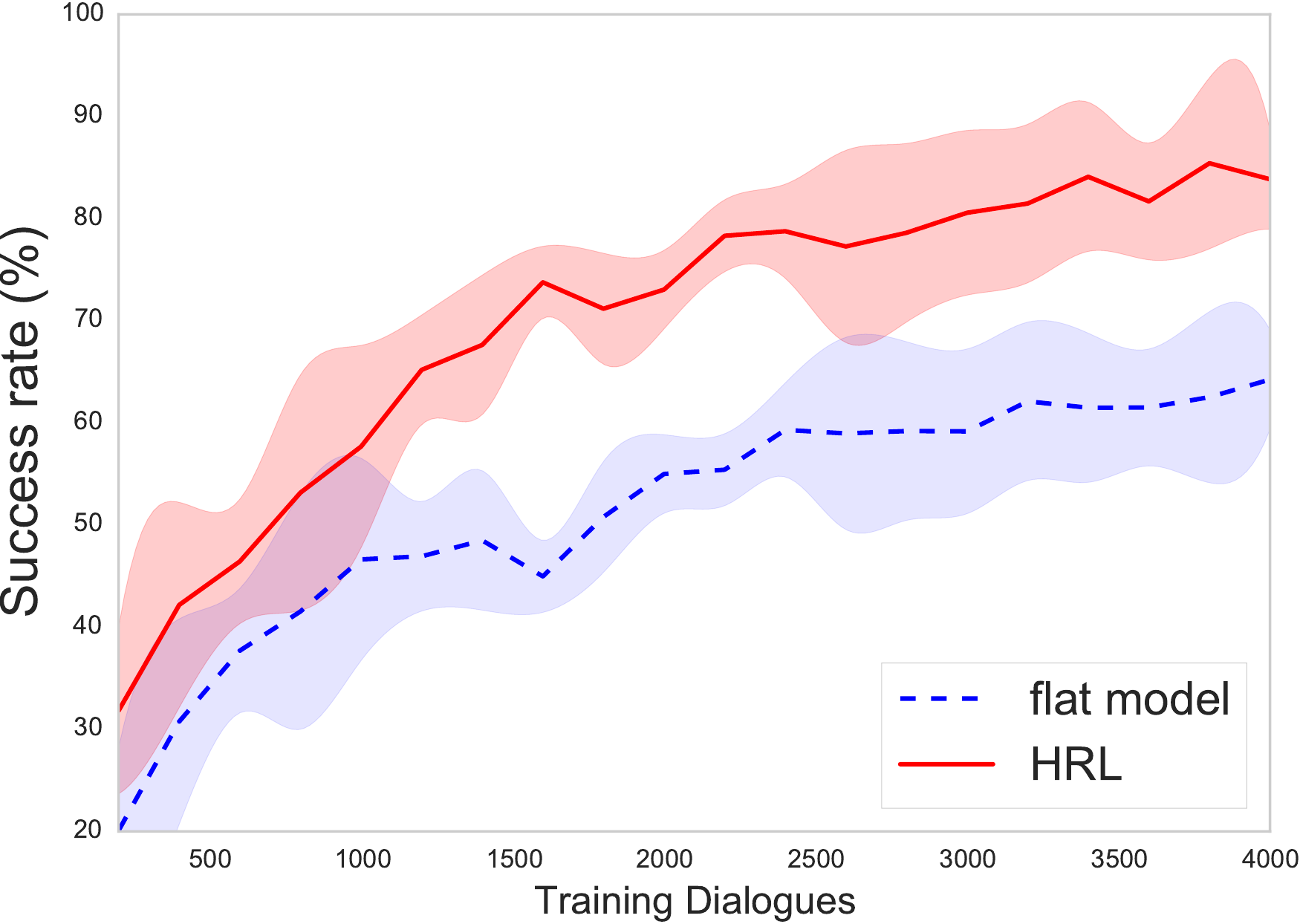}}
\caption{Learning curves for flat and the hierarchical reinforcement learning models.}
\label{fig:ser}
\end{figure}

\section{Experiments}
The PyDial dialogue modelling tool-kit  \cite{ultes2017pydial} 
was used to evaluate the proposed architecture. 
The restaurant domain consists of approximately $100$ venues with $3$ search constraint slots while the hotel domain has $33$ entities with $5$ properties. There are $5$ slots in the booking domain that the system can ask for
while the payment domain has $3$ search constraints  slots. 

In the case of the flat approach, each master domain was combined with the sub-domains, resulting in $11$ and $13$ requestable slots for the restaurants and hotel domains, respectively.

The input for all models was the full belief state $\mathbf{b}$, which expresses the distribution over the user intents and the requestable slots. The belief state has size $311$, $156$, $431$ and $174$ for the restaurants, hotels, booking and payment domains in the hierarchical approach. The flat models have input spaces of sizes $490$ and $333$ for the restaurant and hotel domains accordingly.

The proposed models were evaluated 
with an agenda-based simulated user~\cite{schatzmann2006survey}
where the user intent was perfectly captured in the dialogue belief state.
For both intrinsic and extrinsic evaluation, the total return of each dialogue was set to $\mathds{1}(\mathcal{D})*20- T$, where $T$ is the dialogue length  and $\mathds{1}(\mathcal{D})$ is the success indicator for dialogue $\mathcal{D}$. Maximum dialogue length was set to $30$ in both hierarchical and flat model scenarios with $\gamma = 0.99$.

At the beginning of each dialogue, the master domain is chosen randomly and the user is given a goal which consists of finding an entity and either booking it (for a specific date) or paying for it. The user was allowed to change the goal with a small probability and could not proceed with the sub-domains before achieving the master domain goal. 

\subsection{Hierarchical versus the Flat Approach}
Following \cite{dietterich2000hierarchical,kulkarni2016hierarchical}, we apply a more exploratory policy in the case of master domains, allowing greater flexibility in managing primitive and composite actions during the initial learning stages. Figure \ref{fig:ser} presents the results with $4000$ training dialogues, where the policy was evaluated after each $200$ dialogues.

\begin{figure}[t!]
\centerline{\includegraphics[width=\linewidth]{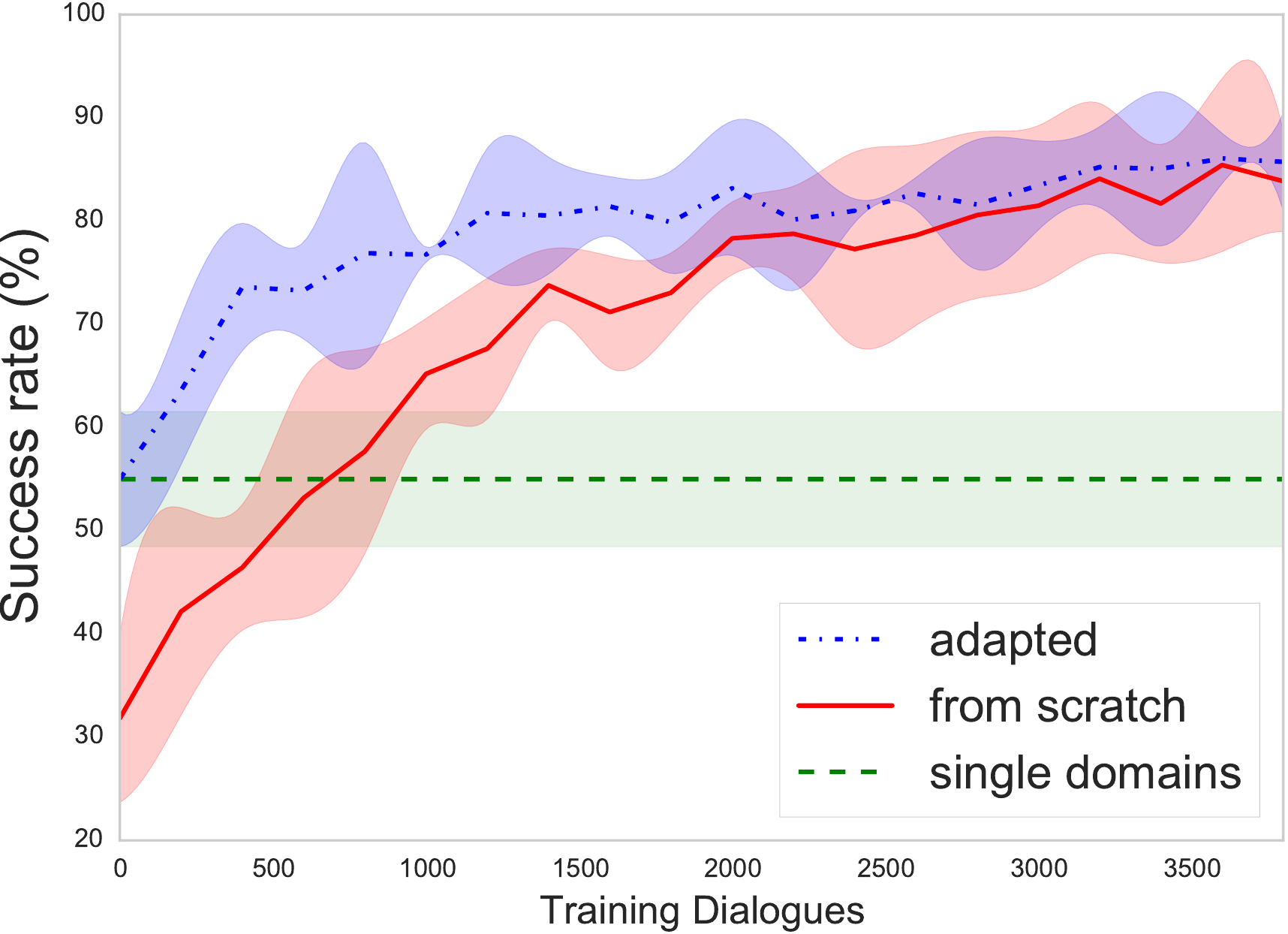}}
\caption{Comparison of policies learnt from scratch and those pre-trained on master domains.}
\label{fig:pretrained}
\end{figure}
The results validate the option framework: it learns faster and leads to a better final policy than the flat approach. The flat model did overcome the cold start problem but it could not match the performance of the hierarchical model.
The policies learnt for sub-tasks with the flat approach perform only $10\%$ worse (on average) than in the hierarchical case. However, providing the entity in both master domains has around $20\%$ lower success rate compared to HRL. 

Moreover, the flat model was not able to match the performance of the HRL approach even with more training dialogues. We let it run for another $6000$ dialogues and did not observe any improvements in success rate (not reported here). This confirms the findings from other RL tasks - the flat approach is not able to remember successful strategies across different tasks \cite{peng2017composite,duan2016benchmarking}. An example of two successful dialogues for both models is presented in the Figure \ref{fig:example}.

\begin{figure}[!ht]
\begingroup
 \fontsize{8.3pt}{9.2pt}\selectfont
\begin{center}
\begin{tabularx}{0.3\textwidth}{l |c}
\toprule
constraint & slot\\     \hline
pricerange & moderate\\    
kind & guesthouse \\     
stars & don't care \\     
hasparking & no \\    
hour &  4 pm\\   
peopleNo &  4\\    
durationDays &  4\\   
day &  don't care \\    
\bottomrule
\multicolumn{2}{c}{requestable slots}\\     \hline
\multicolumn{2}{c}{name of hotel}  \\
\multicolumn{2}{c}{price of hotel}  \\
\end{tabularx}

\begin{tabularx}{\linewidth}{lX}
\toprule
S: & Hello, welcome to the Cambridge Multi-Domain dialogue system. How may I help you? \textbf{[master domain]}\\
U: & Hello. \\
S: & Do you want a hotel or a guesthouse?\\
U: & I'm looking for a guesthouse in a moderate pricerange. I don't care about stars and don't need a parking access.\\
S: & A and B guest house is in a moderate pricerange and has 4 stars. \\
U: & How much does it cost?\\
S: & Cheapest single room is 50 pounds and a cheapest double room is 70 pounds . \\
U: & Ok. Can you book it for 4 people?\\
S: & book() \textbf{[option action]}\\
S: & What time are you planning to check in? \textbf{[sub-domain]}\\
U: & 4 pm.\\
S: & How long do you want to stay?\\
U: & 4 days anytime soon. \\
S: & You want to book it for 4 people, is that right?\\
U: & Yes. \\
S: & I confirm a reservation for 4 days on Saturday for 4 people. Reservation number is 8.  \\
U: & Bye. \\
S: & Goodbye. \textbf{[master domain]}  \\
\bottomrule
\end{tabularx}

\begin{tabularx}{\linewidth}{lX}
\toprule
S: & Hello, welcome to the Cambridge Multi-Domain dialogue system. How may I help you? \\
U: & I am looking for a guesthouse. \\
S: & How many stars should the place have? From 0 to 4.\\
U: & I don't need a parking access. \\
S: & How many stars should the place have? From 0 to 4. \\
U: & I don't care but the price should be moderate.\\
S: & A and B guest house is in a moderate pricerange and has 4 stars. \\
U: & How much does it cost?\\
S: & Cheapest single room is 50 pounds and a cheapest double room is 70 pounds. \\
U: & Ok. Can you book it for 4 people anytime soon?\\
S: & It doesn't have to have a parking, right? \textbf{[wrong domain]}\\
U: & Yes.\\
S: & For what day should I make a reservation?\\
U: & I don't care. \\
S: & You are looking for a 4 stars hotel right? \textbf{[wrong domain]}\\
U: & Yes. \\
S: & Would you like the place to have a parking? \textbf{[wrong domain]}\\
U: & No. \\
S: & How many days do you want to stay there? \\
U: & 4. \\
S: & I confirm a reservation for 4 days on Saturday. Reservation number is 10.  \\
U: & Bye. \\
S: & Goodbye.  \\
\bottomrule
\end{tabularx}
\caption{An example dialogue with the same user goal (top) handled by HRL (middle) and flat (bottom) models.}
\label{fig:example}
\end{center}
\endgroup
\end{figure}

\subsection{Adaptation of Pretrained Policies}
Following the idea of curriculum learning \cite{bengio2009curriculum}, we test the adaptation capabilities of pre-trained policies to more complex situations. Adaptation has proven to be an effective way of reusing existing dialogue policies in new domains \cite{gavsic2014incremental}. Since the kernel  function is factored into the kernel for the belief state space and the action space, we can considered them separately. Following \cite{gavsic2014incremental} the action kernel function is defined only on actions that appear both in original and extended sets and equals $0$ otherwise. The kernel for the belief state space is not changed as we operate on the same belief space.

We first train both master domains (without sub-goals) until robust policies are learned. 
Subsequently, both master domains are re-trained in a hierarchical manner for $4000$ dialogues (testing after each $200$). Figure \ref{fig:pretrained} shows the results compared to the policy learnt from scratch.
Both policies trained on independent domains were able to adapt to more complicated tasks very quickly using the hierarchical framework with new options. This confirms that our approach can substantially speed up learning time by training a policy in a supervised way with the available data and then adapting it to more complex multi-task conversations. 
\section{Conclusion and Future Work}
This paper introduced a hierarchical policy management model for learning dialogue policies which operate over composite tasks. The proposed model uses hierarchical reinforcement learning with the Gaussian Process as the function approximator. Our evaluation showed that our model learns substantially faster and achieves better performance than standard (flat) RL models. 
The natural next step towards the generalisation of this approach is to deepen the hierarchy and compare the performance with deep approaches. Another direction for further work is the validation of this model in interaction with real users.

\section*{Acknowledgments}
Pawe\l{} Budzianowski is supported by EPSRC Council and Toshiba Research Europe Ltd, Cambridge Research Laboratory.

\newpage
\bibliography{acl2017}

\begin{thebibliography}{}
\expandafter\ifx\csname natexlab\endcsname\relax\def\natexlab#1{#1}\fi

\bibitem[{{Bacon} et~al.(2017){Bacon}, {Harb}, and {Precup}}]{bacon2017}
P.-L. {Bacon}, J.~{Harb}, and D.~{Precup}. 2017.
\newblock {The Option-Critic Architecture}.
\newblock {\em 31AAAI Conference On Artificial Intelligence\/} .

\bibitem[{Barto and Mahadevan(2003)}]{barto2003recent}
Andrew~G Barto and Sridhar Mahadevan. 2003.
\newblock Recent advances in hierarchical reinforcement learning.
\newblock {\em Discrete Event Dynamic Systems\/} 13(4):341--379.

\bibitem[{Bengio et~al.(2009)Bengio, Louradour, Collobert, and
  Weston}]{bengio2009curriculum}
Yoshua Bengio, J{\'e}r{\^o}me Louradour, Ronan Collobert, and Jason Weston.
  2009.
\newblock Curriculum learning.
\newblock In {\em Proceedings of the 26th annual international conference on
  machine learning\/}. ACM, pages 41--48.

\bibitem[{Cuay{\'a}huitl(2009)}]{cuayahuitl2009hierarchical}
Heriberto Cuay{\'a}huitl. 2009.
\newblock Hierarchical reinforcement learning for spoken dialogue systems.
\newblock {\em PhD Thesis, University of Edinburgh\/} .

\bibitem[{Cuay{\'a}huitl et~al.(2010)Cuay{\'a}huitl, Renals, Lemon, and
  Shimodaira}]{cuayahuitl2010evaluation}
Heriberto Cuay{\'a}huitl, Steve Renals, Oliver Lemon, and Hiroshi Shimodaira.
  2010.
\newblock Evaluation of a hierarchical reinforcement learning spoken dialogue
  system.
\newblock {\em Computer Speech \& Language\/} 24(2):395--429.

\bibitem[{Cuay{\'a}huitl et~al.(2016)Cuay{\'a}huitl, Yu, Williamson, and
  Carse}]{cuayahuitl2016deep}
Heriberto Cuay{\'a}huitl, Seunghak Yu, Ashley Williamson, and Jacob Carse.
  2016.
\newblock Deep reinforcement learning for multi-domain dialogue systems.
\newblock {\em NIPS Workkshop\/} .

\bibitem[{Dietterich(2000)}]{dietterich2000hierarchical}
Thomas~G Dietterich. 2000.
\newblock Hierarchical reinforcement learning with the maxq value function
  decomposition.
\newblock {\em J. Artif. Intell. Res.(JAIR)\/} 13:227--303.

\bibitem[{Duan et~al.(2016)Duan, Chen, Houthooft, Schulman, and
  Abbeel}]{duan2016benchmarking}
Yan Duan, Xi~Chen, Rein Houthooft, John Schulman, and Pieter Abbeel. 2016.
\newblock Benchmarking deep reinforcement learning for continuous control.
\newblock In {\em Proceedings of The 33rd International Conference on Machine
  Learning\/}. pages 1329--1338.

\bibitem[{Fatemi et~al.(2016)Fatemi, Asri, Schulz, He, and
  Suleman}]{fatemi2016policy}
Mehdi Fatemi, Layla~El Asri, Hannes Schulz, Jing He, and Kaheer Suleman. 2016.
\newblock Policy networks with two-stage training for dialogue systems.
\newblock {\em Proc of SigDial\/} .

\bibitem[{Fikes et~al.(1972)Fikes, Hart, and Nilsson}]{fikes1972learning}
Richard~E Fikes, Peter~E Hart, and Nils~J Nilsson. 1972.
\newblock Learning and executing generalized robot plans.
\newblock {\em Artificial intelligence\/} 3:251--288.

\bibitem[{Ga{\v s}i{\'c} et~al.(2011)Ga{\v s}i{\'c}, Jurcicek, Thomson, Yu, and
  Young}]{milica_real_users}
Milica Ga{\v s}i{\'c}, Filip Jurcicek, Blaise. Thomson, Kai Yu, and Steve
  Young. 2011.
\newblock On-line policy optimisation of spoken dialogue systems via live
  interaction with human subjects.
\newblock In {\em IEEE ASRU\/}.

\bibitem[{Ga{\v{s}}i{\'c} et~al.(2014)Ga{\v{s}}i{\'c}, Kim, Tsiakoulis,
  Breslin, Henderson, Szummer, Thomson, and Young}]{gavsic2014incremental}
Milica Ga{\v{s}}i{\'c}, Dongho Kim, Pirros Tsiakoulis, Catherine Breslin,
  Matthew Henderson, Martin Szummer, Blaise Thomson, and Steve Young. 2014.
\newblock Incremental on-line adaptation of pomdp-based dialogue managers to
  extended domains .

\bibitem[{Ga{\v{s}}i{\'c} et~al.(2015)Ga{\v{s}}i{\'c}, Mrk{\v{s}}i{\'c}, Su,
  Vandyke, Wen, and Young}]{gavsic2015policy}
Milica Ga{\v{s}}i{\'c}, Nikola Mrk{\v{s}}i{\'c}, Pei-hao Su, David Vandyke,
  Tsung-Hsien Wen, and Steve Young. 2015.
\newblock Policy committee for adaptation in multi-domain spoken dialogue
  systems.
\newblock In {\em Automatic Speech Recognition and Understanding (ASRU), 2015
  IEEE Workshop on\/}. IEEE, pages 806--812.

\bibitem[{Ga{\v s}i{\'c} and Young(2014)}]{GPRL}
Milica Ga{\v s}i{\'c} and Steve Young. 2014.
\newblock Gaussian processes for pomdp-based dialogue manager optimization.
\newblock {\em TASLP\/} 22(1):28--40.

\bibitem[{Kulkarni et~al.(2016)Kulkarni, Narasimhan, Saeedi, and
  Tenenbaum}]{kulkarni2016hierarchical}
Tejas~D Kulkarni, Karthik Narasimhan, Ardavan Saeedi, and Josh Tenenbaum. 2016.
\newblock Hierarchical deep reinforcement learning: Integrating temporal
  abstraction and intrinsic motivation.
\newblock In {\em Advances in Neural Information Processing Systems\/}. pages
  3675--3683.

\bibitem[{Laird et~al.(1986)Laird, Rosenbloom, and Newell}]{laird1986chunking}
John~E Laird, Paul~S Rosenbloom, and Allen Newell. 1986.
\newblock Chunking in soar: The anatomy of a general learning mechanism.
\newblock {\em Machine learning\/} 1(1):11--46.

\bibitem[{Lison(2011)}]{lison2011multi}
Pierre Lison. 2011.
\newblock Multi-policy dialogue management.
\newblock In {\em Proceedings of the SIGDIAL 2011 Conference\/}. Association
  for Computational Linguistics, pages 294--300.

\bibitem[{Mnih et~al.(2013)Mnih, Kavukcuoglu, Silver, Graves, Antonoglou,
  Wierstra, and Riedmiller}]{mnih2013playing}
Volodymyr Mnih, Koray Kavukcuoglu, David Silver, Alex Graves, Ioannis
  Antonoglou, Daan Wierstra, and Martin Riedmiller. 2013.
\newblock Playing atari with deep reinforcement learning.
\newblock {\em arXiv preprint arXiv:1312.5602\/} .

\bibitem[{Mrk\v{s}i\'c et~al.(2015)Mrk\v{s}i\'c, {\'O S\'eaghdha}, Thomson,
  Ga\v{s}i\'c, Su, Vandyke, Wen, and Young}]{Mrksic15}
Nikola Mrk\v{s}i\'c, Diarmuid {\'O S\'eaghdha}, Blaise Thomson, Milica
  Ga\v{s}i\'c, Pei-Hao Su, David Vandyke, Tsung-Hsien Wen, and Steve Young.
  2015.
\newblock {Multi-domain Dialog State Tracking using Recurrent Neural Networks}.
\newblock In {\em Proceedings of ACL\/}.

\bibitem[{Parr and Russell(1998)}]{parr1998reinforcement}
Ronald Parr and Stuart~J Russell. 1998.
\newblock Reinforcement learning with hierarchies of machines.
\newblock In {\em Advances in Neural Information Processing Systems\/}. pages
  1043--1049.

\bibitem[{{Peng} et~al.(2017){Peng}, {Li}, {Li}, {Gao}, {Celikyilmaz}, {Lee},
  and {Wong}}]{peng2017composite}
B.~{Peng}, X.~{Li}, L.~{Li}, J.~{Gao}, A.~{Celikyilmaz}, S.~{Lee}, and K.-F.
  {Wong}. 2017.
\newblock {Composite Task-Completion Dialogue System via Hierarchical Deep
  Reinforcement Learning}.
\newblock {\em ArXiv e-prints\/} .

\bibitem[{Schatzmann et~al.(2006)Schatzmann, Weilhammer, Stuttle, and
  Young}]{schatzmann2006survey}
Jost Schatzmann, Karl Weilhammer, Matt Stuttle, and Steve Young. 2006.
\newblock A survey of statistical user simulation techniques for
  reinforcement-learning of dialogue management strategies.
\newblock {\em The knowledge engineering review\/} 21(02):97--126.

\bibitem[{Su et~al.(2017)Su, Budzianowski, Ultes, Ga{\v s}i{\'{c}}, and
  Young}]{eddy2017learning}
Pei-Hao Su, Pawe\l{} Budzianowski, Stefan Ultes, Milica Ga{\v s}i{\'{c}}, and
  Steve~J. Young. 2017.
\newblock Sample-efficient actor-critic reinforcement learning with supervised
  data for dialogue management.
\newblock In {\em Proceedings of the SIGDIAL 2017 Conference\/}.

\bibitem[{Su et~al.(2016)Su, Gasic, Mrksic, Rojas-Barahona, Ultes, Vandyke,
  Wen, and Young}]{su2016continuously}
Pei-Hao Su, Milica Gasic, Nikola Mrksic, Lina Rojas-Barahona, Stefan Ultes,
  David Vandyke, Tsung-Hsien Wen, and Steve Young. 2016.
\newblock Continuously learning neural dialogue management.
\newblock {\em arXiv preprint arXiv:1606.02689\/} .

\bibitem[{Sutton and Barto(1999)}]{RL}
Richard~S. Sutton and Andrew~G. Barto. 1999.
\newblock {\em Reinforcement Learning: An Introduction\/}.
\newblock MIT Press.

\bibitem[{Sutton et~al.(1999{\natexlab{a}})Sutton, McAllester, Singh, Mansour
  et~al.}]{sutton1999policy}
Richard~S Sutton, David~A McAllester, Satinder~P Singh, Yishay Mansour, et~al.
  1999{\natexlab{a}}.
\newblock Policy gradient methods for reinforcement learning with function
  approximation.
\newblock In {\em Proceedings of NIPS\/}. volume~99.

\bibitem[{Sutton et~al.(1999{\natexlab{b}})Sutton, Precup, and
  Singh}]{sutton1999between}
Richard~S Sutton, Doina Precup, and Satinder Singh. 1999{\natexlab{b}}.
\newblock Between mdps and semi-mdps: A framework for temporal abstraction in
  reinforcement learning.
\newblock {\em Artificial intelligence\/} 112(1-2):181--211.

\bibitem[{Ultes et~al.(2017)Ultes, Rojas-Barahona, Su, Vandyke, Kim, Casanueva,
  Budzianowski, Mrk{\v s}i{\'{c}}, Wen, Ga{\v s}i{\'{c}}, and
  Young}]{ultes2017pydial}
Stefan Ultes, Lina~M. Rojas-Barahona, Pei-Hao Su, David Vandyke, Dongho Kim,
  I{\~{n}}igo Casanueva, Pawe{\l} Budzianowski, Nikola Mrk{\v s}i{\'{c}},
  Tsung-Hsien Wen, Milica Ga{\v s}i{\'{c}}, and Steve~J. Young. 2017.
\newblock Pydial: A multi-domain statistical dialogue system toolkit.
\newblock In {\em ACL Demo\/}. Association of Computational Linguistics.

\bibitem[{Wang et~al.(2014)Wang, Chen, Wang, Tian, Wu, and
  Wang}]{wang2014policy}
Zhuoran Wang, Hongliang Chen, Guanchun Wang, Hao Tian, Hua Wu, and Haifeng
  Wang. 2014.
\newblock Policy learning for domain selection in an extensible multi-domain
  spoken dialogue system.
\newblock pages 57--67.

\bibitem[{Williams and Young(2007)}]{POMDP_williams}
Jason~D. Williams and Steve Young. 2007.
\newblock Partially observable {M}arkov decision processes for spoken dialog
  systems.
\newblock {\em Computer Speech and Language\/} 21(2):393--422.

\bibitem[{Williams and Zweig(2016)}]{williams2016end}
Jason~D Williams and Geoffrey Zweig. 2016.
\newblock End-to-end lstm-based dialog control optimized with supervised and
  reinforcement learning.
\newblock {\em arXiv preprint arXiv:1606.01269\/} .

\bibitem[{Young et~al.(2013)Young, Ga{\v{s}}i{\'c}, Thomson, and
  Williams}]{young2013pomdp}
Steve Young, Milica Ga{\v{s}}i{\'c}, Blaise Thomson, and Jason~D Williams.
  2013.
\newblock Pomdp-based statistical spoken dialog systems: A review.
\newblock {\em Proceedings of the IEEE\/} 101(5):1160--1179.

\end{thebibliography}
\bibliographystyle{acl_natbib}
\newpage

\end{document}